\documentclass[10pt,twocolumn,letterpaper]{article}

\usepackage{cvpr}      
\usepackage{siunitx}
\usepackage{multirow}
\usepackage{pifont}
\usepackage{colortbl}
\usepackage{xcolor}
\usepackage{siunitx}
\usepackage{float} 

\PassOptionsToPackage{dvipsnames}{xcolor} 
\usepackage{xcolor}

\usepackage{pifont}
\newcommand{\xmark}{\ding{55}} 

\usepackage[utf8]{inputenc}    
\DeclareUnicodeCharacter{2717}{\xmark} 

\definecolor{cvprblue}{rgb}{0.21,0.49,0.74}
\usepackage[pagebackref,breaklinks,colorlinks,citecolor=cvprblue]{hyperref}

\def\paperID{354} 
\def\confName{3DV\xspace}
\def\confYear{2026\xspace}

\title{Diffusion-Denoised Hyperspectral Gaussian Splatting}

\author{
Sunil Kumar Narayanan \quad
Lingjun Zhao \quad
Lu Gan \quad
Yongsheng Chen\textsuperscript{\dag} \\
Georgia Institute of Technology \\
{\tt\small \{skumar704, lzhao360, lgan\}@gatech.edu \quad yongsheng.chen@ce.gatech.edu}
}

\begin{document}

\maketitle
\let\thefootnote\relax\footnotemark\footnotetext{\textsuperscript{\dag}Corresponding author.}

\begin{abstract}
Hyperspectral imaging (HSI) has been widely used in agricultural applications for non-destructive estimation of plant nutrient composition and precise quantification of sample nutritional elements. Recently, 3D reconstruction methods, such as Neural Radiance Field (NeRF), have been used to create implicit neural representations of HSI scenes. This capability enables the rendering of hyperspectral channel compositions at every spatial location, thereby helping localize the target object's nutrient composition both spatially and spectrally. However, it faces limitations in training time and rendering speed. In this paper, we propose Diffusion-Denoised Hyperspectral Gaussian Splatting (DD-HGS), which enhances the state-of-the-art 3D Gaussian Splatting (3DGS) method with wavelength-aware spherical harmonics, a Kullback–Leibler divergence-based spectral loss, and a diffusion-based denoiser to enable 3D explicit reconstruction of the hyperspectral scenes for the entire spectral range. We present extensive evaluations on diverse real-world hyperspectral scenes from the Hyper-NeRF dataset to show the effectiveness of our DD-HGS. The results demonstrate that DD-HGS achieves the new state-of-the-art performance compared to all the previously published methods. Project page: \href{https://dragonpg2000.github.io/DDHGS-website/}{https://dragonpg2000.github.io/DDHGS-website/}
\end{abstract}

\section{Introduction}

\begin{figure}[t]
  \centering
     \includegraphics[width = \columnwidth]{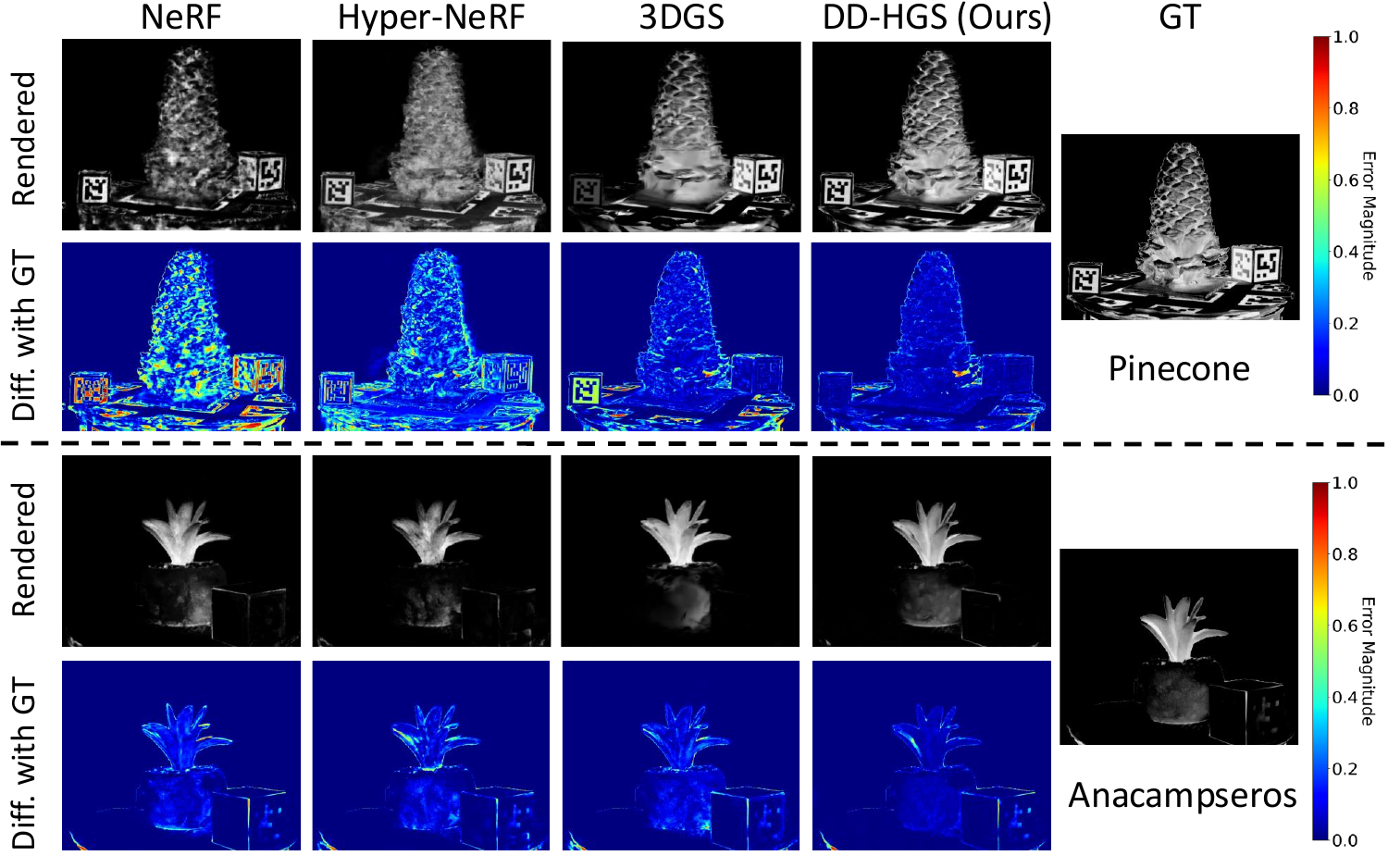}
     \vspace{-15pt}
    \caption{We propose Diffusion-Denoised Hyperspectral Gaussian Splatting (DD-HGS) for reconstructing agricultural scenes and enabling novel view synthesis under hyperspectral imaging. Compared with NeRF~\cite{mildenhall2020nerf}, Hyper-NeRF~\cite{chen2024hyperspectralnerf}, and 3DGS~\cite{kerbl2023gaussians}, ours can render high-quality images with fine-grained spectral details, and significantly reduce reconstruction errors.}
    \vspace{-18pt}
    \label{fig:teaser}
\end{figure}

Human eyes perceive wavelengths in the visible range ($\SI{380}{nm}$–$\SI{750}{nm}$), captured by conventional RGB cameras in three broad channels. In contrast, hyperspectral cameras capture hundreds of narrow wavelength bands across and beyond the visible spectrum, enabling material identification through unique spectral signatures~\citep{spectral_signature}. This makes hyperspectral imaging (HSI) valuable in agriculture, food quality assessment, construction, and environmental monitoring. For example, in agriculture, HSI supports non-destructive nutrient analysis and precise estimation of plant composition~\citep{corti2017hyperspectral,hsinutrient}. Extending HSI to 3D spatial reconstructions enables localization of materials in the physical world, forming the foundation of digital twins—virtual replicas enriched with geometric and spectral properties for predictive and interactive analysis.

Digital twins have shown promise in agriculture for tasks such as mineral analysis, crop yield prediction, fruit counting, and robust detection in low-light conditions~\citep{meyer2024fruitnerf,chopra2024agrinerf}. Integrating spectral information into such models can enhance crop monitoring~\citep{cropmonitoringspace}, yield estimation~\citep{cornyieldhsi,potatoyieldhsi}, and non-destructive nutrient assessment~\citep{HU2021128473,BAI2018492}, ultimately supporting precise agriculture, resource management, and food security under climate and population pressures.

Neural Radiance Fields (NeRFs)~\citep{mildenhall2020nerf} represent scenes using implicit neural networks that map 3D coordinates and viewing directions to color and density, and have become a foundation for 3D reconstruction and novel view synthesis. Extensions of NeRF to hyperspectral data~\citep{chen2024hyperspectralnerf} demonstrate feasibility, but they inherit NeRF’s key limitations, including slow training, high computational cost, and a tendency to overfit sensor noise. These issues are particularly severe for HSI, where narrow spectral bands yield weak signals, making reconstructions highly noise-sensitive~\citep{noisereductionhsi}.

Recently, 3D Gaussian Splatting (3DGS)~\citep{kerbl2023gaussians} replaces implicit neural fields with an explicit representation of the scene as a set of anisotropic 3D Gaussians. Each Gaussian encodes position, orientation, scale, and radiance properties, and images are rendered via fast GPU rasterization. This explicit formulation enables real-time rendering and significantly faster convergence compared to NeRF, while maintaining high-quality geometry and appearance. However, vanilla 3DGS is designed for RGB data and does not account for high-dimensional spectral information or noise robustness.
In this paper, our contributions are as follows:
\begin{itemize}
\item A novel Diffusion-Denoised Hyperspectral Gaussian Splatting (DD-HGS) framework that explicitly models full spectral radiance for 3D hyperspectral reconstruction and novel view synthesis.
\item A wavelength encoder that embeds wavelength-aware spherical harmonics into 3D Gaussians, and a Kullback-Leibler (KL) divergence-based spectral loss to align Gaussian reflectances with ground truth spectra.
\item An integrated hyperspectral diffusion model that denoises 3DGS renderings, improving spectral and spatial fidelity. An end-to-end training paradigm is used.
\item Extensive experiments on diverse real-world hyperspectral scenes from the Hyper-NeRF~\citep{chen2024hyperspectralnerf} dataset demonstrate that DD-HGS achieves the state-of-the-art reconstruction accuracy and rendering quality.
\end{itemize}
\vspace{-1pt}

\section{Related Work}
\subsection{Hyperspectral Imaging}

The main distinction between hyperspectral and RGB imaging lies in spectral resolution. While RGB sensors capture only three broad bands (red, green, and blue), hyperspectral sensors acquire tens to hundreds of narrow, contiguous bands spanning from ultraviolet ($\SI{380}{nm}$) to near-infrared ($\SI{1100}{nm}$)~\citep{ahmad2024comprehensivesurveyhyperspectralimage}. This fine-grained spectral sampling enables detailed material discrimination and has proven useful in agriculture for nutrient assessment~\cite{cohen2025proof}, pollutant detection, and mineral composition analysis.  
However, hyperspectral cameras are highly noise-prone. Narrow-band filters admit fewer photons per band, leading to low signal-to-noise ratios that are further exacerbated by environmental variations and lighting conditions~\citep{liang3dhsmodel,spectral_diffusion,spectral_transformer,Fu_2015_ICCV}. Recent approaches mitigate this by exploiting the strong correlations between adjacent bands that share structural content, using spectral correlation priors~\citep{noisereductionhsi} or low-rank factorization~\citep{lowrankfactorization} to effectively suppress noise and enhance image quality.

\subsection{3D Reconstruction for Hyperspectral Imaging}

NeRF~\citep{mildenhall2020nerf} models volumetric scenes using MLPs, mapping 3D positions and viewing directions to density and color, which are then integrated via volumetric rendering. Hyperspectral extensions include X-NeRF~\citep{poggi2022xnerf} for cross-spectral consistency, Hyperspectral NeRF~\citep{ma2024hyperspectralnerfaug} that outputs $N$ spectral channels, and Hyper-NeRF~\citep{chen2024hyperspectralnerf} with wavelength-aware encoding. While effective, NeRF-based methods are slow to train, expensive to render, and sensitive to hyperspectral noise.  
3D Gaussian Splatting (3DGS)~\citep{kerbl2023gaussians} instead represents scenes using anisotropic Gaussians parameterized by position, scale, orientation, opacity, and spherical harmonics (SH) for view-dependent color, enabling fast optimization and real-time rendering. Numerous variants further extend 3DGS to dynamic scenes~\citep{yang2024deformablegs,wu2024gaussian4d,luiten2024dynamicgs}.
Hyper-GS~\citep{thirgood2025hypergs} adapts Gaussian splatting for hyperspectral data using latent spectral features, with pixel-adaptive density and Gaussian pruning for efficiency. However, compression in a latent space sacrifices spectral fidelity and robustness in noisy conditions. 
Our DD-HGS framework avoids compression by directly modeling full spectral radiance in Gaussians and integrates a hyperspectral diffusion model to denoise renderings, preserving both spatial and spectral fidelity.

\subsection{Hyperspectral Diffusion Models}




Diffusion models~\citep{diffusion} are generative models that learn to reverse a gradual noising process to produce high-fidelity samples. They have recently been applied to hyperspectral enhancement and super-resolution~\citep{wang2024enhancing,cao2024diffusion,cheng2025latent}, treating clean hyperspectral images as outputs of a reverse diffusion process that progressively restores spectral and spatial details. \cite{wang2024enhancing} combines a group-wise autoencoder with diffusion for spectral super-resolution, while \cite{cao2024diffusion} introduces disentangled modulation to preserve spectral and spatial fidelity during sharpening. However, these approaches operate purely in the 2D domain, without modeling 3D geometry or view-dependent effects, making them unsuitable for novel view synthesis or spatially consistent hyperspectral reconstruction.
To address this, we embed a conditional diffusion model into the 3DGS reconstruction pipeline~\citep{kerbl2023gaussians}, denoising and refining rendered hyperspectral images that contain geometric and color artifacts. Unlike prior works such as GaussianObject~\citep{yang2024gaussianobject} and MVSplat360~\citep{chen2024mvsplat360}, which use frozen Stable Diffusion models~\citep{rombach2022latent} trained on large-scale RGB data, we train a hyperspectral diffusion model jointly with 3DGS, explicitly incorporating spectral characteristics for accurate hyperspectral scene reconstruction.


\section{Method}

\begin{figure*}[t!]
\centering
\includegraphics[height = 5 cm,width = 16 cm]{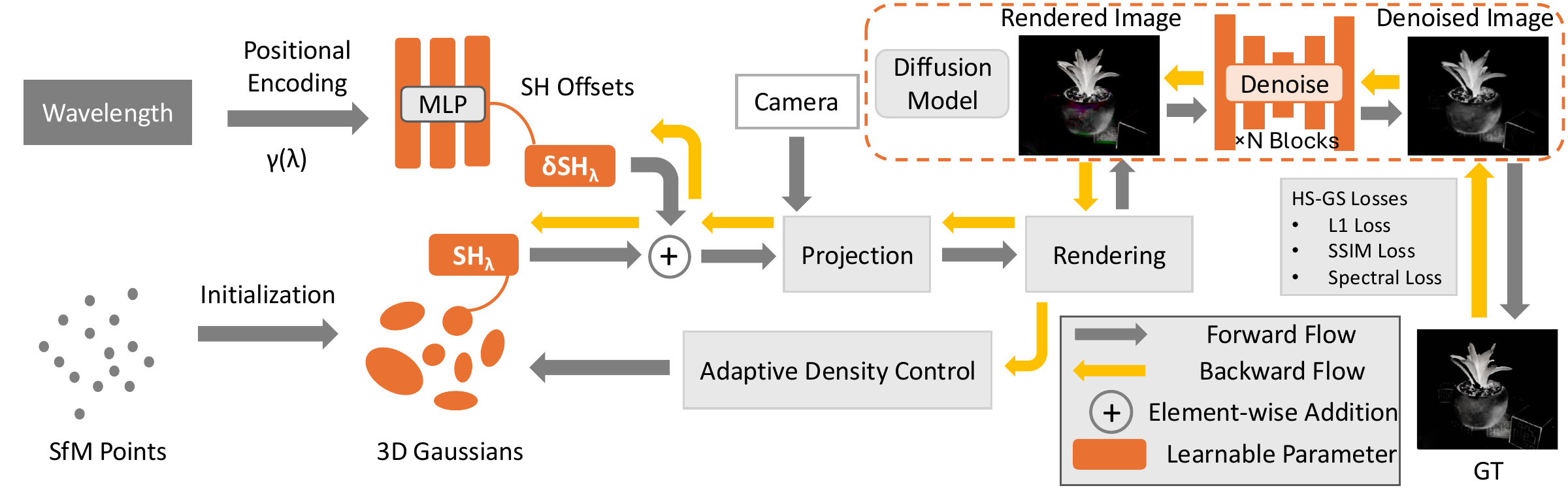}
\vspace{-5pt}
\caption{Overview of our DD-HGS framework. DD-HGS extends 3DGS with a wavelength encoder that maps positional embeddings of wavelength through an MLP to learn wavelength-dependent SH offsets, a spectral loss aligning predicted and ground truth spectral distributions, and a conditional diffusion module that refines the noisy 3DGS rendering to improve its spectral and spatial fidelity.}
\vspace{-10pt}
\label{fig:hsgs}
\end{figure*}

An overview of the proposed Diffusion-Denoised Hyperspectral Gaussian Splatting (DD-HGS) framework is illustrated in Figure~\ref{fig:hsgs}. Our method builds upon 3D Gaussian Splatting~\citep{kerbl2023gaussians} by introducing wavelength-aware modules that enable high-fidelity hyperspectral rendering. We initialize a set of 3D Gaussians from multi-view hyperspectral images with known camera poses, and extend the 3DGS pipeline with the following three key components. First, we introduce a wavelength encoder that maps each input channel's wavelength to a spectral offset through a positional encoding followed by an MLP. These offsets are applied to the spherical harmonics coefficients of the Gaussians to model wavelength-dependent appearance. Secondly, we incorporate a spectral loss that enforces alignment between predicted and ground truth spectral distributions at the pixel level. This loss combines KL divergence and cosine similarity to promote both distributional and angular spectral consistency. Finally, we incorporate a conditional diffusion model to refine the noisy images rendered by 3DGS. This module learns to denoise the output image conditioned on both spatial and spectral context, enhancing fine structures and reducing residual artifacts. Together, these components allow DD-HGS to accurately synthesize spatially-coherent and spectrally-aligned hyperspectral images from sparse multi-view inputs, especially for noisy spectral bands.





\subsection{3D Gaussian Splatting}
3DGS~\citep{kerbl2023gaussians} represents a scene using a set of 3D Gaussians, where each Gaussian is represented by its mean position $\boldsymbol{\mu}$ and covariance matrix 
$\boldsymbol{\Sigma}$:
\begin{align}
     G(\mathbf{x}) = e^{-\frac{1}{2} (\mathbf{x} - \boldsymbol{\mu})^{T} \boldsymbol{\Sigma^{-1}} (\mathbf{x} - \boldsymbol{\mu})},
\end{align}
where $\boldsymbol{\Sigma}$ can be further decomposed into a rotation matrix $\boldsymbol{R}$ and a scaling matrix $\boldsymbol{S}$:
\begin{align}
    \boldsymbol{\Sigma} = \boldsymbol{RSS^TR^T}.
\end{align}
The 3D covariance matrix $\boldsymbol{\Sigma}$ can then be projected onto 2D to enable efficient pixel-wise rendering:
\begin{align}
    \boldsymbol{\Sigma}^{\prime} = \boldsymbol{J W \Sigma W^T J^T},
\end{align}
where $\boldsymbol{\Sigma}^{\prime}$ denotes the 2D covariance matrix, $\boldsymbol{J}$ is the Jacobian of the affine approximation of the projective transformation, and $\boldsymbol{W}$ is the viewing transformation matrix from the world to the camera coordinate frame. 
To render the color of a pixel on the image plane, we use the opacity $\sigma$ and the spherical harmonics (SH) coefficients of the Gaussians to generate 2D views using an $\alpha$-blending algorithm similar to the volumetric rendering in NeRF~\citep{mildenhall2020nerf}.  The rendering process is given as follows:
\begin{align}
    C(&p)   = \sum_{i \in \mathcal{G}}T_i \alpha_i c_i, \\
    T_i = \prod_{j=1}^{i-1}(1-\alpha_{j}) \quad &\text{and} \quad \alpha_i = \sigma_i e^{-\frac{1}{2} (p-\mu_i)^T \Sigma' (p-\mu_i)},
\end{align}
where $C(p)$ denotes the color of a pixel located at $p$, $\mathcal{G}$ is a set of Gaussians along the camera ray sorted by depth with respect to the viewpoint, $T_i$ represents the transmittance, $c_i$ is the color of the Gaussian, and $\mu_{i}$ denotes the 2D coordinate of the 3D Gaussian projected onto the image plane. Detailed projection and rendering processes are described in the original 3DGS paper~\citep{kerbl2023gaussians}.
\subsection{$N$-Channel 3DGS}
To develop a hyperspectral 3DGS framework that can synthesize novel views of hyperspectral images across $N$ different spectral bands, we extend the vanilla 3DGS framework~\citep{kerbl2023gaussians} into an $N$-channel 3DGS ($N$-3DGS) to render images with $N$ wavelength bands. Two main challenges are encountered when extending 3DGS to render multi-channel hyperspectral images. Firstly, traditional structure-from-motion (SfM) methods such as COLMAP~\citep{schoenberger2016sfm} only support grayscale or RGB images. To solve this, we generate pseudo-RGB images from the hyperspectral images using the sensor simulation method in~\cite{chen2024hyperspectralnerf}, and then feed them into COLMAP~\citep{schoenberger2016sfm} to obtain the camera poses and sparse point clouds for Gaussian initialization. Secondly, vanilla 3DGS only supports rendering images with three channels, i.e., red, green, and blue. However, hyperspectral images contain more than three channels, where each channel corresponds to a narrow, specific wavelength band across a continuous spectrum, often covering wavelengths from the visible to near-infrared ranges. Based on the physical properties of hyperspectral imaging, we extend 3DGS to render $N$ spectral channels instead of 3. For each band $i \in \{1,\dots,N\}$, a distinct set of spherical harmonics coefficients ${SH_{i,lm}}$ is associated with each Gaussian. The view-dependent radiance in channel $i$ is then computed as:
\begin{equation}
\vspace{-10pt}
c_i(\mathbf{v}) = \sum_{l=0}^L \sum_{m=-l}^{l} SH_{i,lm} , Y_{lm}(\mathbf{v}),
\vspace{5pt}
\end{equation}
where $\mathbf{v}$ is the view direction and $Y_{lm}(\mathbf{v})$ are the real SH basis functions. This enables each Gaussian to emit reflectance across $N$ wavelength bands, capturing fine-grained spectral variation as a function of view. The per-band radiances are then $\alpha$-blended to generate hyperspectral novel views.

\subsection{Wavelength Encoder}

While 3DGS effectively captures geometry, its SH-based reflectance modeling is suboptimal for hyperspectral data. Without spectral priors, each wavelength channel is treated independently, ignoring the smooth and correlated structure of natural reflectance spectra. This leads to inconsistent SH coefficients and degraded fidelity, particularly at wavelengths outside the visible range.  

To address this, we introduce a wavelength encoder that learns wavelength-specific SH coefficients. Each wavelength $\lambda$ is passed through a positional embedding module with sinusoidal functions at multiple frequencies, enabling the network to capture fine-grained spectral variations. The embedding $\gamma(\lambda)$ is defined as:
\begin{equation}
\gamma(\lambda) = 
\big[\sin(2^k \pi \lambda), \cos(2^k \pi \lambda) \big]_{k=0}^{L-1},
\end{equation}
\noindent where \( L \) denotes the number of frequency bands used in the embedding and thus determines the dimensionality of the resulting positional encoding.

To map these high-frequency features to the same dimension as the SH coefficients, we pass the positional embeddings through a learnable MLP block. The MLP outputs wavelength-specific offsets, which are then added to the SH coefficients for each wavelength band. We use 3D spherical harmonics, where each wavelength is represented by a set of SH coefficients encoding view-dependent color. In our setting, using SH of degree 3 results in 16 coefficients per wavelength. The MLP is designed to match this dimensionality so that the offsets can be added directly to the base SH values. This allows the appearance of each Gaussian to adapt dynamically based on wavelength, improving spectral consistency and rendering quality.
\begin{align}
\delta SH_{\lambda} &= \text{MLP}(\gamma(\lambda)), \\
SH_{\lambda}^{+} &= SH_{\lambda} + \delta SH_{\lambda},
\end{align}
\noindent
where $\delta SH_{\lambda}$ represents the offset of the SH value, $SH_{\lambda}$ and $SH_{\lambda}^{+}$ denote the SH of the 3D Gaussians at the given wavelength $\lambda$ before and after adding the offset, respectively. 
Notably, since other Gaussian parameters, such as position, rotation, and scale, define the intrinsic geometric structure of the scene and are invariant across all wavelengths, these geometric parameters are not modulated.

\subsection{Hyperspectral Denoising with Diffusion Model}
The wavelength encoder captures wavelength-dependent SH coefficients for multi-channel hyperspectral images; however, the rendered outputs can still contain artifacts and noise. To further improve the quality of hyperspectral novel view synthesis, we integrate a diffusion-based denoising model directly into the 3DGS rendering pipeline and train the system end-to-end. Renders from 3DGS serve as inputs to the diffusion model, while its outputs are used to compute the loss for training 3DGS. Unlike conventional post-processing, this design allows the diffusion model to refine hyperspectral renderings during training, thereby improving both denoising and 3DGS reconstruction.

Given an initial render $ \mathbf{X}_{\text{3DGS}} \in \mathbb{R}^{H \times W \times N} $, the ground truth $ \mathbf{X}_{\text{GT}} $ is modeled as the output of a reverse diffusion process conditioned on $ \mathbf{X}_{\text{3DGS}} $. The forward process corrupts $ \mathbf{X}_{\text{GT}} $ with Gaussian noise:
\begin{equation}
\mathbf{X}_t = \sqrt{\bar{\alpha}_t} \mathbf{X}_{\text{GT}} + \sqrt{1 - \bar{\alpha}_t} \boldsymbol{\epsilon}, \quad \boldsymbol{\epsilon} \sim \mathcal{N}(0, \mathbf{I}),
\end{equation}
and the model learns to predict the noise $ \boldsymbol{\epsilon} $ given the noisy input $ \mathbf{X}_{\text{3DGS}} $ by minimizing the following loss:
\begin{equation}
\mathcal{L}_{\text{diffusion}} = \mathbb{E}_{\mathbf{X}_{\text{GT}}, t, \boldsymbol{\epsilon}} \left[ \left\| \boldsymbol{\epsilon} - \boldsymbol{\epsilon}_\theta(\mathbf{X}_t, t \mid \mathbf{X}_{\text{3DGS}}) \right\|^2 \right].
\end{equation}
During training, the diffusion model implicitly learns to correct structured artifacts present in the 3DGS renderings. This joint optimization allows the final outputs to achieve high spectral fidelity while preserving spatial realism. The motivation for using diffusion model stems from the unique challenges of hyperspectral reconstruction: 3DGS renderings exhibit structured, band-specific noise and spectral inconsistencies that simpler models, such as autoencoders, often fail to remove without oversmoothing fine details. Diffusion models, which iteratively refine data, are better suited for eliminating structured noise while preserving high-frequency spectral information. Quantitative comparisons between diffusion and autoencoder baselines are provided on our project page.

\subsection{End-to-End Joint Training}

\subsubsection{Spectral Loss}

Considering the physical properties of hyperspectral images, where each pixel corresponds to a continuous spectral distribution, we argue that to achieve high-quality novel view synthesis, it is essential to ensure that the spectral distributions of the rendered views closely match those of the ground truth. To enforce this alignment, we introduce a spectral loss composed of two terms: a weighted Kullback-Leibler (KL) divergence and a cosine similarity penalty. These jointly encourage the model to produce hyperspectral outputs whose spectral distributions $D_{\lambda}$ accurately reflect those observed in the real scene.

Since KL divergence requires the input to be a probability distribution (i.e., non-negative and summing to 1), we normalize the predicted and ground-truth hyperspectral volumes using a softmax operation along the spectral channel. Specifically, for each pixel $(h, w), h\in H, w \in W$, we define the normalized spectral vector as:
\begin{equation}
\mathbf{D}_{\lambda}(h, w) = \mathrm{softmax}(\mathbf{X}(h, w)),
\end{equation}
where $\mathbf{X} \in \mathbb{R}^{H \times W \times N}$ is the unnormalized hyperspectral volume, and $\mathbf{X}(h, w) \in \mathbb{R}^{N}$ denotes the spectral vector at pixel $(h, w)$. The spectral loss is then computed as:
\begin{equation}
\begin{split}
\mathcal{L}_{\text{spectral}} = & \; \alpha \sum_{h=1}^{H} \sum_{w=1}^{W} 
\mathrm{KL}\!\left( \mathbf{D}_{\lambda}^{\text{GT}}(h,w) \,\|\, 
\mathbf{D}_{\lambda}^{\text{pred}}(h,w) \right) \\
&+ \beta \sum_{h=1}^{H} \sum_{w=1}^{W} 
\Big( 1 - \cos\!\big( \mathbf{D}_{\lambda}^{\text{GT}}(h,w),
\mathbf{D}_{\lambda}^{\text{pred}}(h,w) \big) \Big),
\end{split}
\end{equation}
where $\mathbf{D}_{\lambda}^{\text{GT}}(h,w)$ and $\mathbf{D}_{\lambda}^{\text{pred}}(h,w)$ are the normalized spectral vectors from the ground truth and the prediction at pixel $(h, w)$, respectively.
Here,  $\mathrm{KL}\left( \cdot \,\|\ \cdot \right)$ denotes the Kullback-Leibler divergence, which penalizes distributional mismatches between the predicted and ground truth spectra, while $\cos\left( \cdot, \cdot \right)$ denotes the cosine similarity, which promotes angular alignment between the spectral vectors. The weights $\alpha$ and $\beta$ are user-defined hyperparameters controlling the relative importance of each term in our experiments.
This spectral-aware formulation encourages the network to match not only the absolute intensities but also the shape and directionality of the spectral profiles, which is crucial for downstream applications like material classification and reflectance estimation.
\subsubsection{Overall Loss Function}



To enable end-to-end joint training, we combine the vanilla 3DGS loss terms with the designed spectral and diffusion losses to form the final loss \( \mathcal{L}_{\text{DD-HGS}} \), as shown below:
\begin{equation}
\mathcal{L}_{\text{DD-HGS}} = w_1 \mathcal{L}_{\text{L1}} + w_2 \mathcal{L}_{\text{SSIM}} + w_3 \mathcal{L}_{\text{spectral}} + w_4 \mathcal{L}_{\text{diffusion}}.
\end{equation}
In this equation, \( \mathcal{L}_{\text{L1}} \) measures the pixel-wise difference between the rendered and ground truth images, and \( \mathcal{L}_{\text{SSIM}} \)  evaluates their structural similarity using the Structural Similarity Index Measure (SSIM). The spectral loss \( \mathcal{L}_{\text{spectral}} \) enforces pixel-wise consistency between the predicted and ground truth spectral distributions. Finally, \( \mathcal{L}_{\text{diffusion}} \) supervises the training of the diffusion model, which takes the noisy 3DGS render as input and refines it toward the clean hyperspectral target, correcting geometric distortions and spectral noise. The weights \( w_1 \), \( w_2 \), \( w_3 \), and \( w_4 \) are hyperparameters that balance the contributions of each term.

\label{sec:results}
\section{Datasets}{

We evaluate DD-HGS on two hyperspectral agricultural datasets introduced in Hyper-NeRF~\citep{chen2024hyperspectralnerf}, collected using two distinct imaging systems. Following the benchmark, we split each scene into 90\% training and 10\% testing images.

\textbf{BaySpec Dataset.} Captured with a BaySpec GoldenEye snapshot sensor, this dataset provides hyperspectral images at a spatial resolution of $640 \times 512$ across 141 narrow spectral bands spanning \SI{400}{nm}–\SI{1100}{nm}. The sensor captures full spectral cubes in a single exposure. The scenes consist of three artificial plant species: \textit{Anacampseros}, \textit{Caladium}, and \textit{Pinecone}, mounted on a motorized turntable and imaged at \SI{20}{cm}. Each scene contains 433 hyperspectral views captured over a wide range of viewing angles.

\textbf{Surface Optics Dataset.} Acquired using a Surface Optics SOC710-VP imaging spectrometer, this dataset includes 128 spectral channels spanning \SI{370}{nm}–\SI{1100}{nm}, with a spatial resolution of $696 \times 520$. The sensor uses a line-scanning (pushbroom) mechanism to capture spectral cubes. The camera’s narrow field of view and shallow depth of field necessitated a fixed tripod setup \SI{2}{m} from the subject. Two artificial plants, \textit{Rosemary} and \textit{Basil}, were placed on a rotating stage inside a Macbeth SpectraLight light booth to ensure consistent illumination. Each scene is captured from 48 distinct viewpoints.

}
\section{Quantitative Results}{
We evaluate DD-HGS on in total five plant scenes from the aforementioned two datasets. The implementation details and evaluation metrics are provided on our project website.
We compare our method against a carefully selected group of baselines including NeRF-based and 3DGS-based methods, each chosen for their relevance to hyperspectral novel view synthesis. 
NeRF~\citep{mildenhall2020nerf} and 3DGS~\citep{kerbl2023gaussians} serve as foundational models for radiance field rendering and explicit 3D representation, respectively. 
Hyper-NeRF~\citep{chen2024hyperspectralnerf} is selected for its ability to model hyperspectral reflectances with spectral priors. 
MipNeRF~\citep{barron2021mipnerf} and MipNeRF360~\citep{barron2022mipnerf360} are included due to their strong performance in novel view synthesis tasks. Both leverage mipmapping and hierarchical sampling to effectively handle aliasing and unbounded scene geometry, making them robust candidates for high-fidelity view generation. 
TensoRF~\citep{chen2022tensorrf} is selected for its efficiency and compactness via tensor decomposition, which benefits rendering of high-dimensional outputs. 
Hyper-GS~\citep{thirgood2025hypergs}, a recent extension of 3DGS to hyperspectral rendering, represents the most competitive prior tailored specifically for this task. For a fair comparison, all baselines are extended to support $N$-channel hyperspectral outputs, allowing direct evaluation of both spatial reconstruction quality and spectral fidelity.

\begin{table}[t]
  \centering
  \small
  \setlength{\tabcolsep}{4pt}
  \begin{tabular}{lccccc}
    \toprule
    Method & PSNR$\uparrow$ & SSIM$\uparrow$ & SAM$\downarrow$ & RMSE$\downarrow$ & FPS$\uparrow$ \\
    \midrule
    NeRF~\citep{mildenhall2020nerf}       & 23.35 & 0.606 & 0.0440 & 0.0687 & 0.13 \\
    MipNeRF~\citep{barron2021mipnerf}    & 22.75 & 0.594 & 0.0435 & 0.0776 & 0.09 \\
    TensoRF~\citep{chen2022tensorrf}    & 24.66 & 0.648 & 0.0501 & 0.0587 & 0.17 \\
    Nerfacto~\citep{tancik2023nerfstudio}   & 19.12 & 0.586 & 0.0551 & 0.1174 & 0.50 \\
    MipNeRF360~\citep{barron2022mipnerf360} & 26.53 & 0.744 & \underline{0.0280} & 0.0476 & 0.01 \\
    Hyper-NeRF~\citep{chen2024hyperspectralnerf} & 19.82 & 0.671 & 0.0534 & 0.1071 & 0.47 \\
    3DGS~\citep{kerbl2023gaussians}       & 22.91 & 0.632 & 0.0468 & 0.0810 & \textbf{78.10} \\
    Hyper-GS~\citep{thirgood2025hypergs}   & \underline{27.11} & \underline{0.780} & \textbf{0.0254} & \underline{0.0440} & 2.31 \\
    \textbf{DD-HGS (Ours)} & \textbf{27.18} & \textbf{0.940} & 0.0348 & \textbf{0.0347} & \underline{2.43} \\
    \bottomrule
  \end{tabular}
  \vspace{-5pt}
  \caption{\textbf{Quantitative results on the BaySpec dataset (averaged over Pinecone, Caladium and Anacampseros).} Our method achieves the new state-of-the-art performance among all the published work. Best results in \textbf{bold}, second-best \underline{underlined}.}
  \vspace{-15pt}
  \label{tab:bayspec_results}
\end{table}
\textbf{BaySpec Results.}
Table~\ref{tab:bayspec_results} summarizes results averaged over Pinecone, Caladium, and Anacampseros scenes captured with the BaySpec GoldenEye camera. DD-HGS achieves the best PSNR, SSIM, and RMSE, significantly surpassing prior methods. While Hyper-GS attains a slightly better SAM, our method achieves superior perceptual quality and spectral fidelity. More importantly, DD-HGS maintains a rendering speed of 2.43 FPS, outperforming Hyper-GS in both performance and efficiency. These results highlight that DD-HGS not only advances spectral and spatial reconstruction quality, but also delivers competitive efficiency compared to prior NeRF and 3DGS baselines.

\textbf{Surface Optics Results.}
Table~\ref{tab:surface_optics_part1} reports results on the Rosemary and Basil scenes from the Surface Optics dataset. DD-HGS achieves a PSNR of 38.34, far surpassing Hyper-GS and 3DGS, and reducing RMSE and SAM to 0.003. These substantial gains highlight the effectiveness of diffusion-based refinement in correcting structural errors and residual spectral noise. Baseline approaches such as 3DGS and Hyper-GS struggle with geometric artifacts in regions with high occlusion or sharp spectral transitions, whereas DD-HGS achieves reconstructions with both higher geometric fidelity and precise spectral consistency. Overall, these results establish DD-HGS as the new state-of-the-art method for 3D reconstruction and novel view synthesis on challenging hyperspectral agricultural scenes.
\begin{table}[t]
  \centering
  \small
  \setlength{\tabcolsep}{4pt}
  \begin{tabular}{@{}lccccc@{}}
    \toprule
    Method & PSNR$\uparrow$ & SSIM$\uparrow$ & SAM$\downarrow$ & RMSE$\downarrow$ & FPS$\uparrow$ \\
    \midrule
    NeRF~\cite{mildenhall2020nerf}         &  9.17 & 0.650  & 0.054 & 0.441 & 0.13 \\
    MipNeRF~\cite{barron2021mipnerf}       & 12.33 & 0.578  & 0.54  & 0.371 & 0.09 \\
    TensoRF~\cite{chen2022tensorrf}        & 13.67 & 0.657  & 0.033 & 0.315 & 0.20 \\
    Nerfacto~\cite{tancik2023nerfstudio}   & 17.60 & 0.814  & 0.044 & 0.134 & 0.57 \\
    MipNeRF360~\cite{barron2022mipnerf360} & 11.20 & 0.805  & 0.069 & 0.293 & 0.01 \\
    Hyper-NeRF~\cite{chen2024hyperspectralnerf}
                                           & 17.71 & 0.829  & 0.012 & 0.139 & 0.49 \\
    3DGS~\cite{kerbl2023gaussians}         & 23.38 & \underline{0.954}  & 0.006 & 0.072 & \textbf{79.00} \\
    Hyper-GS~\cite{thirgood2025hypergs}    & \underline{26.04} & \textbf{0.967} & \underline{0.004} & \underline{0.051} & \underline{3.56} \\
    \textbf{DD-HGS (Ours)}                 & \textbf{38.34} & 0.927  & \textbf{0.003} & \textbf{0.003} & 2.95 \\
    \bottomrule
  \end{tabular}
  \vspace{-5pt}
  \caption{\textbf{Quantitative results on the Surface Optics dataset (averaged over Rosemary and Basil).} 
  DD-HGS achieves large gains in PSNR and RMSE, while matching or surpassing prior methods in spectral metrics. 
  Best results in \textbf{bold}, second-best \underline{underlined}.}
  \vspace{-13pt}
  \label{tab:surface_optics_part1}
\end{table}

\begin{figure*}[t]
  \centering
  \includegraphics[height = 12 cm,width = 14 cm]{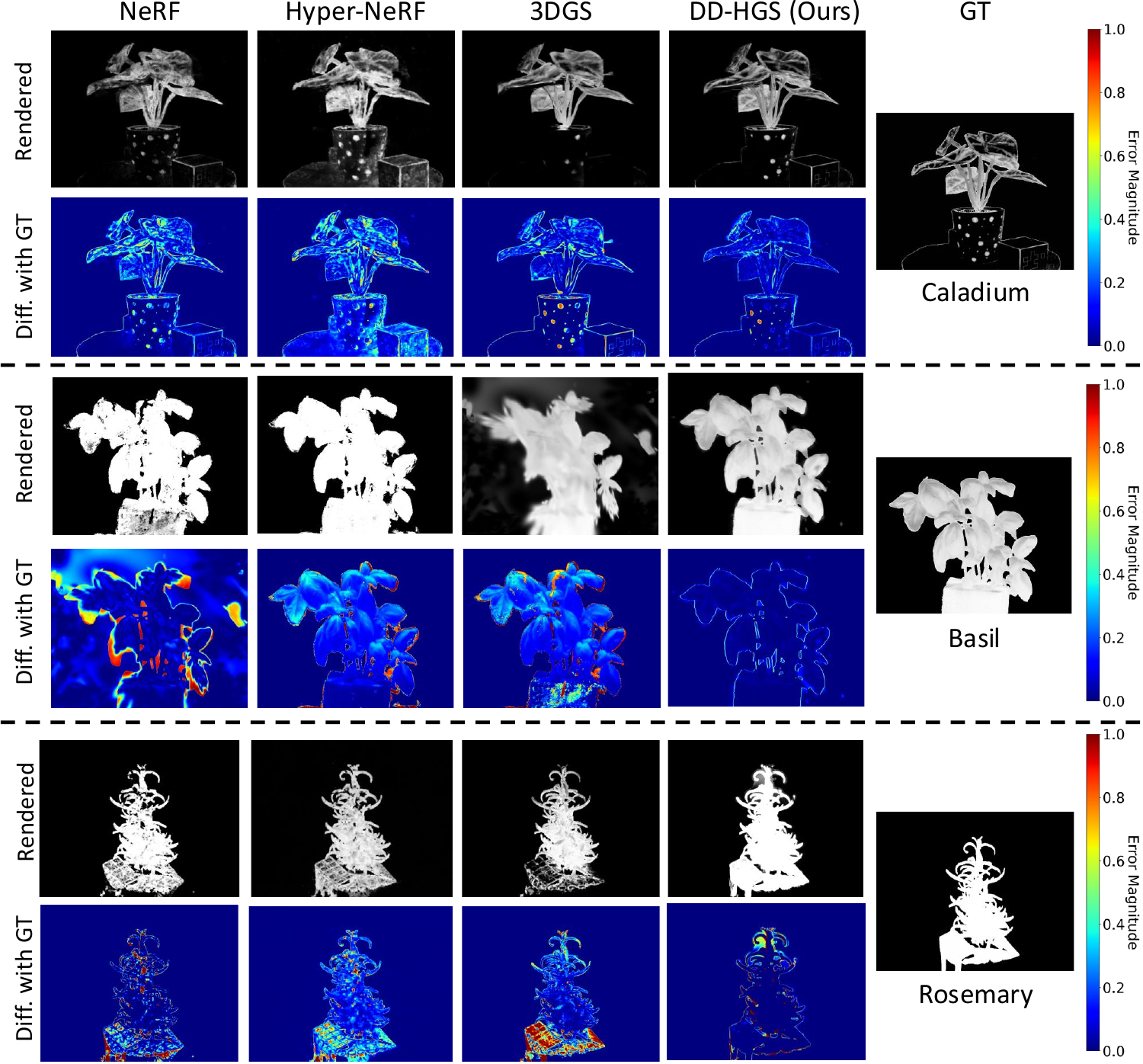}
  \vspace{-5pt}
  \caption{\textbf{Qualitative comparisons on BaySpec and Surface Optics datasets (750–768 nm).} Each column shows renderings and difference heatmaps for NeRF, Hyper-NeRF, 3DGS, and DD-HGS. Our method preserves fine structural details and spectral fidelity.}
  \vspace{-10pt}
  \label{fig:main_qualitative}
\end{figure*}
}
\section{Qualitative Results}{

In Figure~\ref{fig:main_qualitative}, we present rendered hyperspectral images and difference heatmaps against the ground truth of our method and the baselines for three hyperspectral scenes: Caladium, Basil and Rosemary. We qualitatively compare our method with NeRF~\citep{mildenhall2020nerf}, Hyper-NeRF~\citep{chen2024hyperspectralnerf}, and 3DGS~\citep{kerbl2023gaussians}. 
We omit qualitative comparisons with Hyper-GS~\citep{thirgood2025hypergs}, given that its implementation is not publicly available.
NeRF and Hyper-NeRF fail to recover fine-grained spatial details and suffer from large spectral reconstruction errors. 3DGS achieves finer structural reconstruction but still exhibits spectral inconsistencies. In contrast, our DD-HGS significantly reduces reconstruction artifacts and achieves higher geometric accuracy across both hyperspectral camera systems.

\begin{figure*}[t]
  \centering
    \includegraphics[height = 11 cm,width = 15 cm]{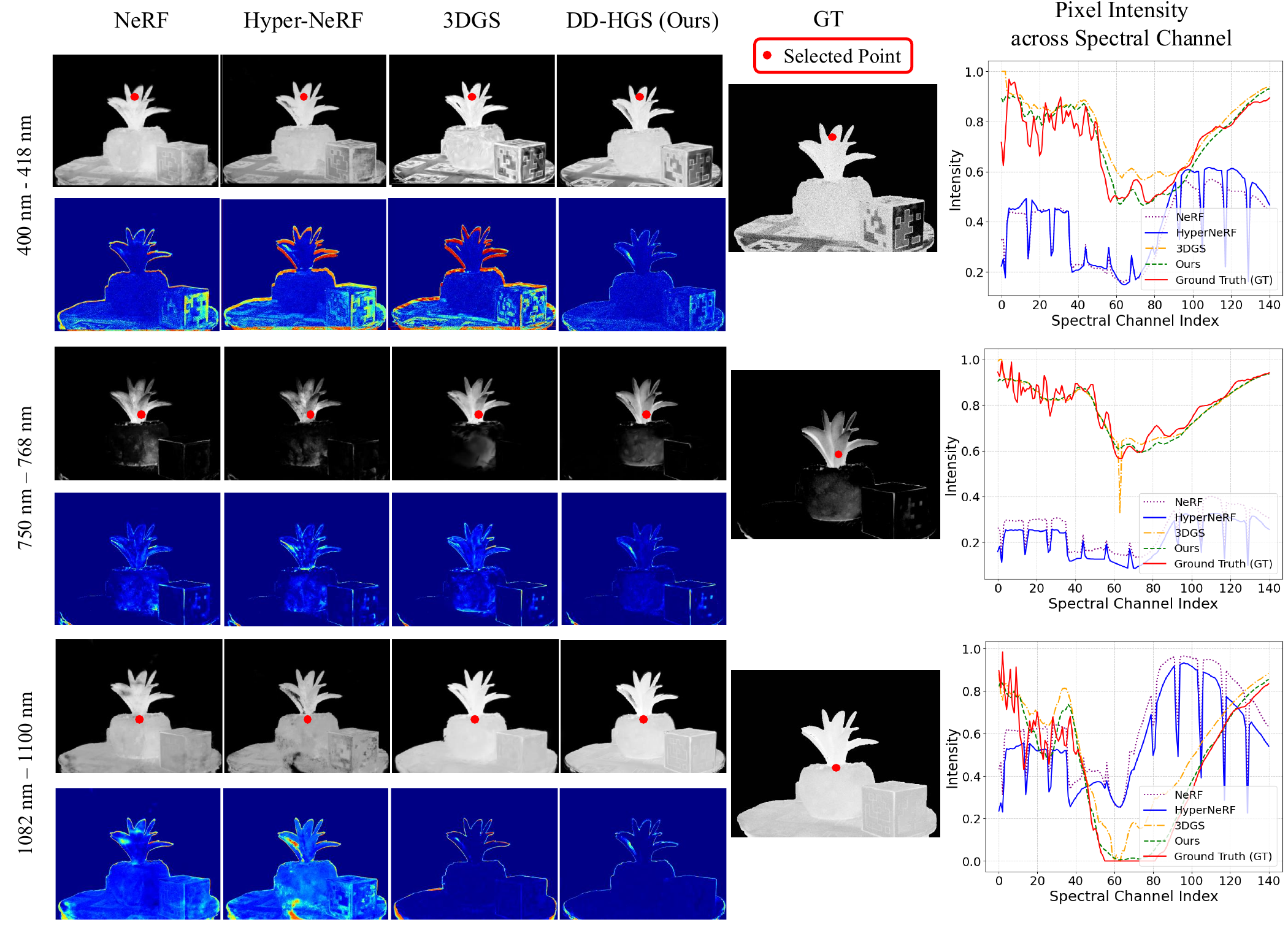}
    \vspace{-5pt}
  \caption{\textbf{Qualitative results on the Anacampseros scene across three different wavelength ranges:} \SI{400}{nm} to \SI{418}{nm}, \SI{750}{nm} to \SI{768}{nm}, \SI{1082}{nm} to \SI{1100}{nm}. The rendered images and difference heatmaps against the ground truth demonstrate the spectral fidelity and spatial consistency of the reconstruction results, particularly under challenging near-infrared and ultraviolet conditions. In addition, we visualize the reconstructed pixel intensities across all the spectral channels of three randomly selected points in the rightmost column. Compared to the baselines, our method exhibits the highest similarity to the ground truth. 
  }
    \vspace{-5pt}
  \label{fig:wavelength_comparison}
\end{figure*}

\begin{figure*}
  \centering
  \includegraphics[height = 5 cm,width = 15 cm]{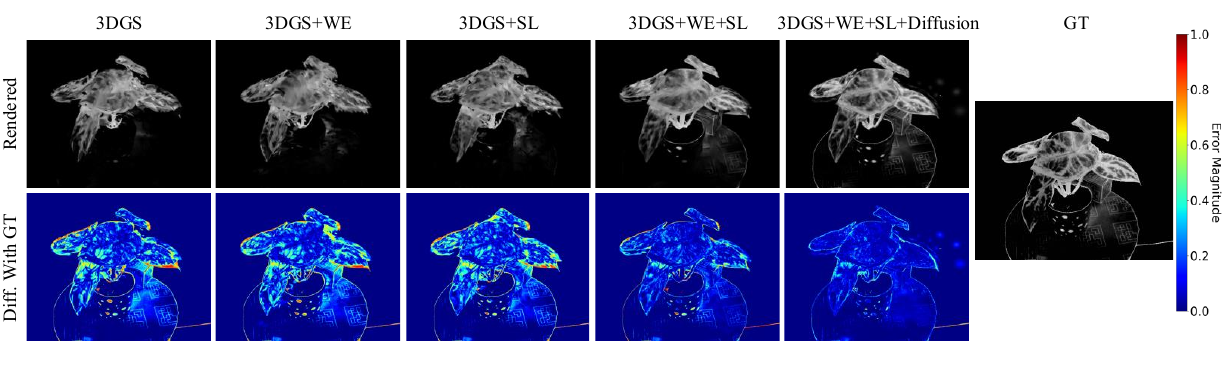}
  \vspace{-18pt}
  \caption{\textbf{Qualitative ablation study on the Caladium scene.} Rendered images and difference heatmaps w.r.t.\ ground truth are shown. The wavelength encoder (WE) and spectral loss (SL) progressively reduce detail artifacts and spectral distortions, leading to higher spatial and spectral reconstruction accuracy.}
  \vspace{-10pt}
  \label{fig:qual_comparison_ablation}
\end{figure*}

To evaluate the generalization of our model across spectral bands, Figure~\ref{fig:wavelength_comparison} shows results on the Anacampseros scene for three ranges: 400-418 nm (ultraviolet), 750-768 nm (near-infrared), and 1082-1100 nm (far-infrared). NeRF~\citep{mildenhall2020nerf} exhibits large deviations in spectral curves due to overfitting to noise, leading to unstable reconstructions. Hyper-NeRF~\citep{chen2024hyperspectralnerf} alleviates this in the near-infrared but still suffers from spectral misalignment. 3DGS~\citep{kerbl2023gaussians} achieves smoother curves yet fails to capture fine spectral transitions. In contrast, DD-HGS consistently aligns with ground truth, producing accurate spatial and spectral reconstructions across all wavelength ranges. Difference heatmaps show that NeRF and Hyper-NeRF generate high-magnitude, spatially inconsistent errors, especially in low-signal ultraviolet and far-infrared bands, while 3DGS yields smoother but broadly distributed residuals. DD-HGS exhibits concentrated, low-magnitude errors localized near geometric edges, reflecting both spatial precision and spectral fidelity. The pixel intensity curves (rightmost column) further demonstrate that DD-HGS (green dashed) closely matches the ground truth (red solid), outperforming all baselines. More visualizations are available on our website.


 }
\section{Ablation Study}
 {We assess the contribution of each module in our DD-HGS through an ablation study in Table~\ref{tab:ablation_results}. The wavelength encoder improves the modeling of wavelength-dependent variations by injecting high-frequency spectral cues into the 3D Gaussian representation. The spectral loss further constrains reconstructions to match ground-truth spectra, reducing spectral misalignment and improving detail recovery. Their combination produces finer-grained structures with consistent spectral fidelity. Adding the diffusion module enables iterative refinement of geometry and spectra, effectively denoising structured artifacts and stabilizing sharp spectral transitions. A “3DGS + Diffusion” baseline further confirms that diffusion alone enhances spatial consistency but still underperforms the full pipeline due to the lack of wavelength-aware supervision. As shown in Fig.~\ref{fig:qual_comparison_ablation}, distortions are progressively reduced as more modules are introduced, culminating in state-of-the-art results with the full DD-HGS method. 
\begin{table}[t]
  \centering
  \setlength{\tabcolsep}{4pt}
  \resizebox{\columnwidth}{!}{%
  \begin{tabular}{@{}lccc|cccc@{}}
    \toprule
    Method & SL & WE & Diff. & PSNR~$\uparrow$ & SSIM~$\uparrow$ & SAM~$\downarrow$ & RMSE~$\downarrow$ \\
    \midrule
    3DGS              & ✗ & ✗ & ✗ & 21.47 & 0.8280 & 0.0649 & 0.0646 \\
    3DGS + SL         & \checkmark & ✗ & ✗ & 21.69 & 0.8286 & 0.0554 & 0.0588 \\
    3DGS + WE         & ✗ & \checkmark & ✗ & 21.52 & 0.8279 & 0.0532 & 0.0576 \\
    3DGS + WE + SL    & \checkmark & \checkmark & ✗ & 21.96 & 0.8313 & 0.0494 & 0.0530 \\
    3DGS + Diffusion  & ✗ & ✗ & \checkmark & 26.17 & 0.9316 & 0.0378 & 0.0383 \\
    \textbf{DD-HGS (Ours)} & \checkmark & \checkmark & \checkmark & \textbf{27.18} & \textbf{0.9400} & \textbf{0.0348} & \textbf{0.0347} \\
    \bottomrule
  \end{tabular}
  }
  \vspace{-5pt}
  \caption{\textbf{Ablation study of main components.} Average results across Pinecone, Anacampseros, and Caladium. Each component improves reconstruction, with DD-HGS (ours) achieving the best.}
  \vspace{-10pt}
  \label{tab:ablation_results}
\end{table}
 }
\section{Conclusion}
\label{sec:conclusion}
In this work, we introduce Diffusion-Denoised Hyperspectral Gaussian Splatting (DD-HGS), a framework for hyperspectral 3D reconstruction and novel view synthesis. Our method integrates a wavelength encoder for spectral conditioning and a spectral loss for alignment with ground truth spectral distributions. To further improve spectral fidelity, we embed a diffusion-based denoiser into the 3DGS rendering pipeline, refining intermediate hyperspectral renders through a conditional reverse diffusion process. The joint optimization corrects structured artifacts such as band-wise noise and geometric distortions, yielding noise-resilient, high-quality reconstructions. Extensive experiments show that DD-HGS consistently outperforms prior methods in both spectral accuracy and spatial \mbox{realism}.
\textbf{Limitations: }Our proposed method has been validated exclusively on scenes with a single object. However, due to the lack of large-scale open-source multi-view hyperspectral imaging datasets, our model has not been evaluated on scenes with multiple, diverse objects. Future work will extend our method to hyperspectral 3D reconstruction of multi-object scenes in complex environments.\\

\noindent
\textbf{Acknowledgments.} 
Sunil Kumar Narayanan was supported in part by the U.S. Department of Agriculture (Award Nos. 2024-67021-42876, 2024-67021-43862, and 2024-67021-41534) and the National Science Foundation (Award Nos. 2112533, 2345543, and 2419122). 
Lingjun Zhao was supported by the IRIM Ph.D. Fellowship at the Georgia Institute of Technology.


\small
\bibliographystyle{ieeenat_fullname}

\end{document}